\documentclass[runningheads]{llncs}

\usepackage{eccv}
\usepackage{ulem}

\usepackage{eccvabbrv}
\usepackage{multirow}
\usepackage{graphicx}
\usepackage{booktabs}
\usepackage[accsupp]{axessibility}  
\usepackage{colortbl}  
\usepackage{xcolor}
\usepackage{array}   
\definecolor{mygray}{gray}{.9}

\usepackage[pagebackref,breaklinks,colorlinks,citecolor=eccvblue]{hyperref}

\usepackage{orcidlink}

\begin{document}

\title{CrossGLG: LLM Guides One-shot Skeleton-based 3D Action Recognition in a Cross-level Manner} 

\titlerunning{CrossGLG}

\author{Tingbing Yan\inst{1} \and
Wenzheng Zeng\inst{1\dag} \and
Yang Xiao\inst{1\dag}   \and
Xingyu Tong\inst{1} \and
Bo Tan\inst{1} \and
Zhiwen Fang\inst{2,3} \and
Zhiguo Cao\inst{1} \and
Joey Tianyi Zhou\inst{4,5}} 

\authorrunning{Yan et al.}

\institute{Key Laboratory of Image Processing and Intelligent Control, Ministry of Education, School of Artificial
Intelligence and Automation, Huazhong University of Science and Technology, Wuhan 430074, China \and
School of Biomedical Engineering, Southern Medical University, Guangzhou 510515, China \and
Department of Rehabilitation Medicine, Zhujiang Hospital, Southern Medical University, Guangzhou 510280, China \and
Centre for Frontier AI Research, Agency for Science, Technology and Research (A*STAR), Singapore \and
Institute of High Performance Computing, Agency for Science, Technology and Research (A*STAR), Singapore 
\email{\{yantingbing,wenzhengzeng,Yang\_Xiao,xy\_tong,bo\_tan,zgcao\}@hust.edu.cn},\\fzw310@smu.edu.cn, zhouty@cfar.a-star.edu.sg}

\maketitle
\let\thefootnote\relax\footnotetext{\dag~Wenzheng Zeng and Yang Xiao are corresponding authors.}

\begin{abstract}
  Most existing one-shot skeleton-based action recognition focuses on raw low-level information (e.g., joint location), and may suffer from local information loss and low generalization ability. To alleviate these, we propose to leverage text description generated from large language models (LLM) that contain high-level human knowledge, to guide feature learning, in a global-local-global way. Particularly, during training, we design $2$ prompts to gain global and local text descriptions of each action from an LLM. We first utilize the global text description to guide the skeleton encoder focus on informative joints (i.e.,global-to-local). Then we build non-local interaction between local text and joint features, to form the final global representation (i.e., local-to-global). To mitigate the asymmetry issue between the training and inference phases, we further design a dual-branch architecture that allows the model to perform novel class inference without any text input, also making the additional inference cost neglectable compared with the base skeleton encoder. Extensive experiments on three different benchmarks show that CrossGLG consistently outperforms the existing SOTA methods with large margins, and the inference cost (model size) is only $2.8$\% than the previous SOTA. CrossGLG can also serve as a plug-and-play module that can substantially enhance the performance of different SOTA skeleton encoders with a neglectable cost during inference. The source code will be released soon.


\end{abstract}

\section{Introduction}
\label{sec:intro}

Driven by the accessibility of low-cost 3D cameras like the Microsoft Kinect~\cite{zhang2012microsoft}, 3D skeleton-based action recognition has emerged as an active area of research. Despite the remarkable progress, most works~\cite{yan2018spatial,plizzari2021spatial,li2019actional,lee2023hierarchically}  focus on fully supervised settings that heavily rely on large-scale annotated data for feature learning, introducing high annotation costs. One feasible way to alleviate this is to conduct one-shot 3D action feature learning.

To this end, some pioneer~\cite{ALCAGCN,motionbert,skeletondml,SLDML,uDTW,ntu120} attempts have been made. They are all based on the raw 3D skeleton data, focusing on low-level information (e.g., joint localization).  While they tend to suffer from the following defeats: 
(1) Existing methods generally can not focus on the crucial local areas, which leads to the loss of important details.
(2) The lack of high-level semantic information to guide the models makes it difficult to grasp the deep overall motion characteristics, resulting in a weak generalization for unseen actions. 

Some studies~\cite{percetionofmotion,motionpercetionsurvey} in the fields of psychology and neuroscience show that humans can easily identify critical motion clues and extrapolate local observations into global conclusions to recognize actions, even with a limited amount of observations.
Inspired by this, our motivation is to utilize high-level human knowledge to guide feature learning, for effective one-shot action recognition. Thanks to the recent success of large language models trained on massive web data~\cite{GPT3,GPTpre_train,InstructGPT,thought_chain}, we can manage to obtain high-level human knowledge from the text they generate. Thus, we propose CrossGLG, a novel architecture that utilizes knowledgeable text descriptions to guide skeleton feature learning in a global-local-global way. 

\begin{figure}[t]
  \centering
   \includegraphics[width=0.85\linewidth]{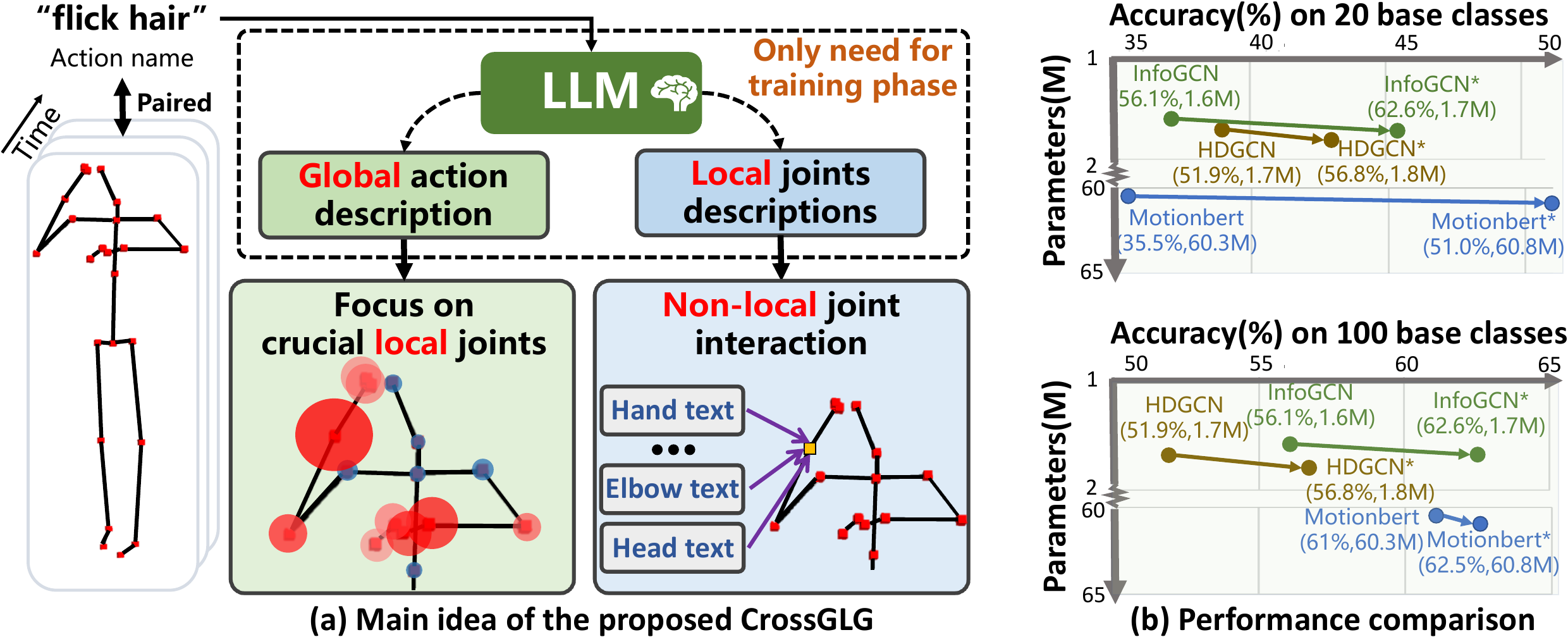}

   \caption{(a) Main idea of the proposed CrossGLG: We propose to leverage text description generated from large language models (LLM) that contain high-level human knowledge to guide
feature learning, in a global-local-global way. In global-to-local (block in green), the larger the radius of the circle around a joint, the more important that joint is. In local-to-global (block in blue), non-local interaction establishes connections between all textual features and all skeleton features at the joint level to summarize the high-level global action representation. (b) Performance comparison on NTU RGB+D 120~\cite{ntu120} dataset with 20 and 100 base classes: CrossGLG can serve as a plug-and-play module that can substantially enhance the performance of different SOTA skeleton encoders with a neglectable cost during inference..}
   \label{fig:fig1}
   \vspace{-5pt}
\end{figure}

Particularly, as shown in Fig.~\ref{fig:fig1}(a), the paradigm can be divided into $2$ main phases: global-to-local (block in green), and local-to-global (block in blue). We first utilize a global action description to guide the skeleton encoder to focus on local informative joints (i.e., global-to-local). The global action description is obtained from an LLM by a designed prompt, which illustrates the important joints when a specific action occurs. We regard those mentioned joints as informative joints and design a Joint Importance Determination Module (JID) to let the skeleton encoder focus on those informative local joints. Such global-to-local guidance can let the skeleton encoder focus more on the important local clues for effective motion representation.

Based on the enhanced local joint features, we further build a non-local interaction between local joint skeleton features and local joint-level motion descriptions generated from the LLM, to achieve the so-called local-to-global. Within it, effective feature exchange between local joint-level text-text, skeleton-skeleton, and skeleton-text has been established to enhance the global-aware ability of local features, forming a global summary of the action. In conclusion, the proposed cross-modal global-local-global guiding strategy lets the skeleton encoder first identify critical motion clues and then extrapolate local observations into global conclusions to effectively recognize actions.

Another issue is that for practical applications, the network is not capable of using text data during inference. To mitigate the asymmetry issue between the training and inference phases, we design a dual-branch architecture. Specifically, we divide the whole framework into a skeleton encoding branch and a cross-modal guidance branch, where the skeleton encoding branch is only responsible for skeleton feature encoding, and the cross-modal guidance branch is responsible for guiding the feature learning of the skeleton encoding branch, in the aforementioned global-local-global way. During training, we introduce a shared classifier to those two branches to reduce the feature gap between them. Such design not only lets features in the skeleton branch learn from the cross-modal guidance branch during training, but also keeps those two-branch features consistent so that only the skeleton encoder is needed during inference, without the need for any additional modal data.

The proposed CrossGLG makes the feature encoder learn fine-grained high-level semantic information from the text description both globally and locally. 
As shown in \cref{fig:fig1} (a), in the action `flick hair', the joints of the arm and hand are of greater importance than joints such as the spine. Meanwhile, our visualization results show that for the novel class action that was not seen during training, the model without fine-tuning can still pay more attention to the informative clues. 
This implies that the model's spatial information grasping ability has been greatly improved and can really focus on important local regions with strong generalization ability. Extensive experiments on three benchmark (i.e., NTU RGB+D 60~\cite{ntu60}, NTU RGB+D 120~\cite{ntu120}, and Kinetics~\cite{kinetics}) also validate the effectiveness of the proposed method. Our method achieves state-of-the-art performance on both datasets in terms of accuracy and efficiency. Notably, CrossGLG outperforms the SOTA methods by $6.6\%$ and $5.9\%$ in the experimental settings of 20 and 80 base classes on NTU RGB+D 120~\cite{ntu120}, while maintaining a model size that is only 2.8\% of the previous SOTA~\cite{motionbert}. Moreover, as shown in~\cref{fig:fig1} (b), CrossGLG can also serve as a plug-and-play module that can substantially enhance the performance of different SOTA skeleton encoders~\cite{motionbert,HDGCN, infogcn} with a neglectable cost during inference.

In summary, our contributions are summarized below:

$\bullet$ We are the first to consider text descriptions from large language models to facilitate one-shot skeleton-based 3D human action recognition.

$\bullet$ We design CrossGLG, a novel architecture that utilizes knowledgeable text descriptions to guide skeleton feature learning in a global-local-global way, for effective one-shot 3D action recognition.

$\bullet$ We propose a dual-branch architecture to address the asymmetry issue between training and testing.

$\bullet$ CrossGLG can serve as a plug-and-play module that can substantially enhance the performance of different SOTA skeleton encoders with a neglectable cost during inference.

\section{Related Work}
\textbf{One-shot 3D skeleton-based action recognition. } To alleviate the labor-consuming issues in 3D action recognition, one-shot skeleton-based action recognition has emerged as a research focal point and some efforts~\cite{meta,Lifeifei} have been paid.
These works focus on enhancing the adaptation ability of feature extractor and classifier from base class to novel class via imitating the one-shot test cases using base class data during training. There is also a portion of work~\cite{motionbert,skeletondml,SLDML,uDTW,ntu120} that opts for another approach, matrix learning techniques. They enhance the discriminative properties of generated features by exploring the construction of latent feature space, which enlarges the distance of features between different classes and narrows the distance of features for the same class of actions. 

However, these works only focus on the low-level information of the skeleton sequence (i.e., joint locations). Due to the lack of guidance from high-level semantic information, these methods face problems such as low generalization ability. Although APSR~\cite{ntu120} has tried to introduce semantic information, this introduction is too little. And it cannot realize the detection of important joints during inference, which limits the improvement of the discriminative properties of the test features. Unlike these previous approaches, we utilize human-knowledge-rich action description texts to introduce high-level semantic information to guide the learning of skeleton features from both global and local perspectives.


\label{sec:relate_work}

\noindent\textbf{Skeleton action recognition with cross-modal information.}
Some existing works also utilize auxiliary information to facilitate skeleton feature learning, which is mainly based on two types of additional inputs: visual data~\cite{2streamRGB1,2streamRGB3,2streamRGB4,2streamdepth1,2streamdepth2,2streamRGBco1,2streamRGBco2} and non-visual data~\cite{GAP}.
The first class mainly utilizes RGB or depth data. They design a two-stream network to process visual data and skeleton data separately and perform score fusion, feature fusion, or co-learning. However, these visual data may suffer from the problems of expensive acquisition, noise, and inefficient semantic knowledge. In the second category, GAP~\cite{GAP} performs contrastive learning between skeleton features and text features to facilitate feature learning. Despite the effectiveness, it treats all local features equally, which reduces the discriminative power. Besides, it simply conducts contrastive learning within each local-level and global-level skeleton-text feature pair, lacking the cross and non-local interactions between different parts of the body, which limits the global-aware ability of local features. 
Moreover, it is designed for fully supervised setting, and the experiment (\cref{tab:NTU120_compare_other}) demonstrates its inferior generalization performance in the more challenging one-shot setting. Suitable design under the challenging one-shot learning paradigm has not been well concerned yet. To our knowledge, we are the first to consider knowledgeable text descriptions to guide skeleton feature learning for one-shot 3D action recognition, with proper and effective designs. We propose a novel cross-modal guidance framework that runs in a global-local-global way. Experiments demonstrate the significant superiority of our novel design.


\noindent\textbf{Large Language Model (LLM).}
Benefiting from unique training mechanisms~\cite{GPTpre_train,GPT_context,GPT_RLHF,thought_chain,InstructGPT,GPT_train}, massive parameters~\cite{GPT3,GPT1}, and sufficient training data~\cite{GPT_data2}, large language models (e.g., ChatGPT) typically can understand a wider and more complex language context. This enables them to better understand the semantics and contextual relevance in the text and maintain strong text generation capabilities. That is, given the corresponding prompt, LLMs can generate logical and detailed results. 
Inspired by this, we propose to utilize the generative power of large language models to assist us in extracting global and local strongly semantic heuristic information about actions.
\begin{figure*}[t]
  \centering
   \includegraphics[width=0.99\linewidth]{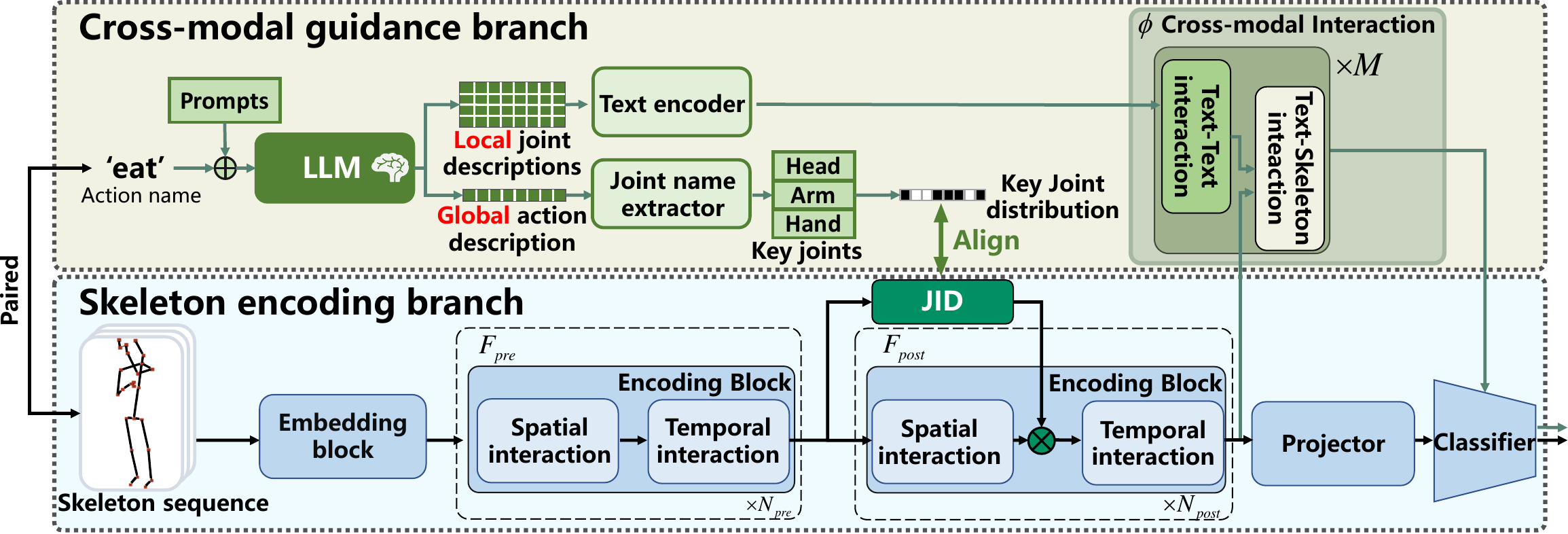}

   \caption{The overall model architecture. JID denotes the Joint Importance Discrimination module that outputs the importance of each joint based on the skeleton features. We design a cross-modal guidance branch (colored with green) to guide the skeleton feature learning. During novel class inference, only the skeleton encoding branch (colored with blue) will be used without any textual input.}
   \label{fig:pipeline}
\end{figure*}
\section{Method}
\label{sec:method}
Here we will introduce the proposed CrossGLG, a novel cross-modal guided one-shot skeleton-based human action recognition architecture. We will briefly recap the task formulation of one-shot action recognition in Sec.~\ref{subsec:preliminary}. The architecture overview will be illustrated in Sec.~\ref{subsec:architecture}, followed by detailed descriptions of each component (from Sec.~\ref{subsec:txt_collection} to Sec.~\ref{subsec:overall_scheme}).

\subsection{Preliminary}
\label{subsec:preliminary}
In one-shot 3D action recognition, we are given a labeled base class dataset $\mathcal{D}_{base}$ and a novel class dataset $\mathcal{D}_{novel}$, where the two datasets do not overlap. $\mathcal{D}_{novel}$ consists of N action classes, where one sample of each class along with its label is drawn to form the support set $\mathcal{S}_{novel}$, and the others to form the query set $\mathcal{Q}_{novel}$. The goal is to train the model using the $\mathcal{D}_{base}$ to complete the training of the model and make adaptations on $\mathcal{S}_{novel}$ to make predictions for $\mathcal{Q}_{novel}$.

\subsection{Architecture Overview}
\label{subsec:architecture}
The architecture of the proposed CrossGLG is shown in \cref{fig:pipeline}, which contains two branches: the skeleton encoding branch and the cross-modal guidance branch. The skeleton encoding branch takes a skeleton sequence as input, encodes it, and ultimately performs action classification. We utilize the proposed cross-modal guidance branch to guide the skeleton feature learning. Specifically, based on the paired action name of the input skeleton sequence, we design two kinds of prompts that allow an LLM (e.g., ChatGPT) to generate both global action description, and local joint-level motion description of the action (Sec.~\ref{subsec:txt_collection}). 
We use the obtained global action description to guide the skeleton encoder focusing on local informative joints, by aligning the output distribution between the designed Joint Importance Determination Module (JID) and the key joints distribution derived from the global action description. 
We further introduce a cross-modal interaction module $\phi$ to build strong non-local interactions between local joint features and joint-level text features, and summarize the high-level global information for the entire action. Further detailed illustrations can be found in Sec.~\ref{subsec:GLG}. The two-branch outputs will share the same classifier for the final action classification. During inference, only the skeleton encoding branch will be used to conduct one-shot novel class classification, without using any auxiliary text information. 

\subsection{Derive Knowledgeable Action Descriptions from Large Language Model (LLM)}
\label{subsec:txt_collection}
\begin{figure}[t]
  \centering
   \includegraphics[width=1\linewidth]{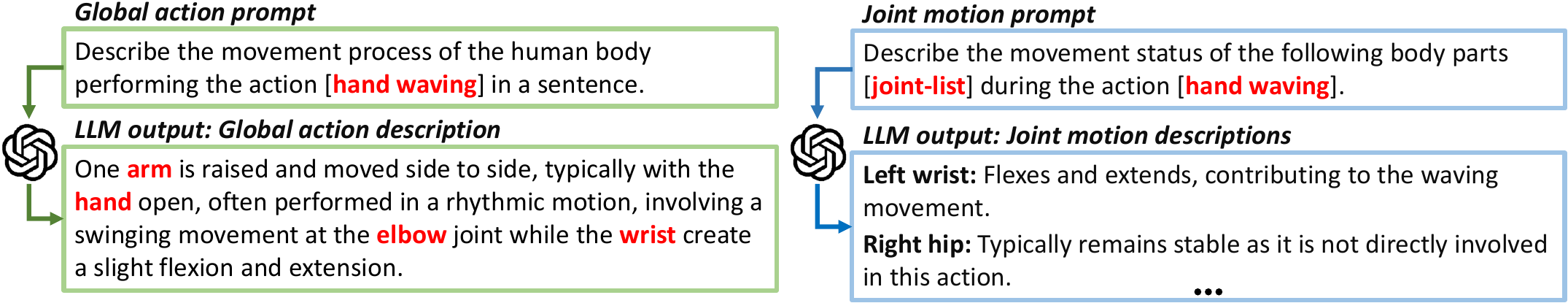}

   \caption{The paradigm to acquire knowledgeable action description texts through a large language model (ChatGPT~\cite{GPT3} is used in our implementation).}
   \label{fig:text_form}
   \vspace{-15pt}
\end{figure}

Our motivation is to utilize high-level human knowledge to guide the skeleton feature learning, for a more effective one-shot action recognition. Text is actually a good carrier to involve such knowledge, as it is low-noise and rich in high-level semantic information. Thanks to the recent success of Large Language Models (LLM)~\cite{GPT3,GPTpre_train,InstructGPT,thought_chain} trained on massive web data, we can manage to obtain high-level human knowledge from the text they generate. 
To this end, we designed two prompts (i.e., global action prompt and joint motion prompt) to obtain both global and local joint-level descriptions of the action, as shown in \cref{fig:text_form}. 
The global description of the specified action and its joint motion description can be obtained from the LLM by simply changing the `action name' (e.g., `[hand waving]' in the given example) and `[joint-list]' (e.g., the defined $25$ joints in the NTU RGB+D 60 dataset~\cite{ntu60}) in the global action prompt and joint motion prompt. The global action description indicates which joints and body parts are important from a global perspective. The joint-level motion description provides fine-grained high-level semantic information from a local perspective.
The two types of text will be subsequently used to guide the skeleton feature learning.

\subsection{Cross-Modal Global-Lobal-Global Guidance} \label{subsec:GLG}

We propose to guide the feature learning from $2$ aspects: (1) using global action description to guide the skeleton encoder focus on local informative joints (i.e., global-to-local guidance), and (2) establishing non-local interactions between local joint-level text and skeleton features, to achieve a kind of local-to-global cross-modal guidance.

\subsubsection{Global-to-Local Textual Guidance}
\label{subsubsec:G2L}
\textbf{Obtaining local informative joints from the global action description text.} 
As shown in \cref{fig:text_form}, the acquired global action description indicates which areas are required to accomplish a particular movement from a global perspective (i.e., the word in red color). Those informative regions are actually more essential to recognize the action. Therefore, our motivation is to extract those key areas from the global action description and guide the feature encoder to pay more attention to them.
Particularly, we utilize the Noun Phrase Extraction module of Stanford CoreNLP~\cite{coreNLP} to extract the nouns in the global action description. 
Among these nouns, only joints and body parts such as `arm' are left. Moreover, the body parts are converted to the corresponding joints in the human skeleton defined by the dataset used. For example, in the NTU RGB+D 120 dataset~\cite{ntu120}, arms correspond to shoulders, elbows, and wrists, and eyes correspond to the head. Thus, we can get the distribution of key joints from the global description of each action. Let $K \in \mathbb{R}^{V}$ be the distribution matrix of key joints of an action, where $K_{j}=1$ if the $j$-th is a key joint of the action. Here, $V$ represents the number of joints.

\noindent\textbf{Joint Importance Determination Module (JID).}
We design JID to let the skeleton encoder focus on the extracted informative joints. As shown in ~\cref{fig:pipeline}, 
the base skeleton encoder (e.g., InfoGCN~\cite{infogcn} and MotionBERT~\cite{motionbert}) will typically contain an embedding block, $N$ encoding blocks, a projector, and a classifier.
Each encoding block performs a spatial interaction and a temporal interaction on the skeleton features. Spatial interaction refers to the interaction of different joint features at the same time, while temporal interaction refers to the interaction of the same joint feature at different times. We divide the $N$ encoding blocks into two parts: The first part $F_{pre}$ has $N_{pre}$ blocks and the second part $F_{post}$ has $N_{post}$ blocks. 
After processing the input 3D skeleton sequence $X \in \mathbb{R}^{T_{in}\times V\times C_{in}}$ by the embedding block, we first feed it into $F_{pre}$ to obtain a skeleton feature $f_{pre} \in \mathbb{R}^{T_{pre}\times V\times C_{pre}}$ that gets processed with certain spatio-temporal information. 
Then we obtain a preliminary joint overall motion feature $\overline{f}_{pre}\in \mathbb{R}^{V\times C_{pre}}$ by pooling $f_{pre}$ along the time dimension. Here, $T_{in}$, $T_{pre}$ denote the sequence length. $C_{in}$, $C_{pre}$ denote the channel lengths of the input and $f_{pre}$. We feed the $\overline{f}_{pre}$ into the Joint Importance Discriminator (JID) module to obtain the importance of each joint motion feature, where JID is composed of two linear layers and a softmax layer. JID receives the $\overline{f}_{pre}$ and outputs the joint importance $k_{out}\in \mathbb{R}^{V}$ of $X$, where $k_{out}^i$ represents the importance of the $i$-th joint and it holds that $0\le k_{out}^i\le 1$ for all $1\leq i \leq V$. $F_{post}$ receives $f_{pre}$ to generate $f_{post}\in \mathcal{R}^{T\times V\times C_{post}}$. The generated $k_{out}\in \mathcal{R}^{V}$ is used to reweight the spatially interacted feature $f_{post}$ in each encoding block of $F_{post}$ as show in ~\cref{eq:multi},
\begin{equation}
  \left[ f_{post} \right] _{ij}^{t}= k_{out}^i \cdot \left[ f_{post} \right] _{ij}^{t}, t\in \left[ 1,T \right] , i\in \left[ 1,V \right] , j\in \left[ 1,C_{post} \right],
  \label{eq:multi}
\end{equation}
where i and j iterate over the joint and channel dimensions respectively. This can help the encoder focus more on local informative joints for effective action representation.
To make JID produce more reasonable results, we design a calibration loss to align its output to the key joints distribution $k_{gt}\in \mathbb{R}^{V}$ that is derived from the global action text description:

\begin{equation}
  \mathcal{L}_{calibrate} = MSE(k_{out}, k_{gt}),
  \label{eq:key_joints}
\end{equation}

where MSE is the standard MSE-loss. By aligning $k_{out}$ to $k_{gt}$, JID can benefit from the high-level knowledge from the global action description to output a more reasonable joint importance result.

\subsubsection{Local-to-Global Cross-Modal Interaction.}
\label{subsubsec:L2G}
Although the skeleton modality provides simple and efficient information about the body structure, it may also suffer from sparse representation and potential noise~\cite{ARsurvey}.
If the model only focuses on the motion of local skeleton joints, it will be difficult to capture the high-level overall motion features, to summarize as an action-level representation. This may reduce the effectiveness of feature representation and hurt the generalization capability, especially under the one-shot setting. Therefore, we further design a local-to-global cross-modal interaction module $\phi$. Within it, non-local interaction between joint-level motion description and joint-level skeleton features is built to summarize the high-level global representation of the entire action.    

Specifically, we use a pre-trained text encoder (e.g., DeBerta~\cite{deberta} in our implementation) to process the joint-level motion descriptions (illustrated in \cref{subsec:txt_collection}) to get their feature embeddings $t \in \mathbb{R}^{V\times C_{txt}} $. 
We pass the skeleton sequence $X$ through the skeleton encoding blocks $F_{pre}$ and $F_{post}$ to get the feature $f_{post} \in \mathbb{R}^{T \times V\times C_{post}}$, and pool $f_{post}$ along the time dimension to get the overall joint features $\overline{f}_{post} \in \mathbb{R}^{V\times C_{post}}$. We project $t$ and $\overline{f}_{post}$ to a public feature space $P$ by 2 learnable MLPs, resulting in $p_{txt}\in \mathbb{R}^{V\times C_{p}}$ and $p_{ske}\in \mathbb{R}^{V\times C_{p}}$, respectively. Here, $C_{txt}$ and $C_{post}$ are respectively the channel lengths of the text feature space and skeleton space, and $C_p$ is the channel length of the public feature space. After that, we feed $p_{txt}$ and $p_{ske}$ into $\phi$ for interaction. $\phi$ has a total of M layers of interaction blocks, where $\phi ^i$ is the $i$-th block in $\phi$ ($1\leq i \leq M$). Each of these blocks has the same architecture. In each block $\phi^ i$, we build non-local interactions in text-text, skeleton-skeleton, and skeleton-text perspectives.

First, we put $p_{txt}$ through a self-attention module to allow each joint text feature to get information from other text features of other joints. As shown in \cref{eq:self-attention}, $p_{txt}$ is fused with the non-local semantic context information, resulting in $p_{txt}^{i}\in \mathbb{R}^{V\times C_{p}}$:
\begin{equation}
  p_{txt}^{i} = MHA(Q,K,V=p_{txt}),
  \label{eq:self-attention}
\end{equation} 
where MHA~\cite{transformer} is the multi-head attention.
We then utilize text feature $p_{txt}^{i}$ to guide the integration of joint-level skeleton features $p_{st}^{i-1}$ to obtain $p_{st}^{i}\in \mathbb{R}^{V\times C_{p}}$ through a cross-attention module, as shown in \cref{eq:cross-attention}, 
\begin{equation}
  p^i_{st} = MHA(Q = p_{txt}^{i}, K = p_{st}^{i-1}, V = p_{st}^{i-1}),    
  \label{eq:cross-attention}
\end{equation} 
where $p_{st}^{0}$ is $p_{ske}$.

After that, we sum up the text features $p_{txt}^{i}$ and skeleton features $p_{st}^{i}$ to further fuse the knowledge of the two modal data through MLP processing, as shown in \cref{eq:res_connection},
\begin{equation}
  p^i_{st} = MLP(p_{txt}^{i}+p^i_{st}).    
  \label{eq:res_connection}
\end{equation} 
After going through the fusion operation of $M$-layer blocks, the output $p_{st}^{M} \in \mathbb{R}^{V\times C_{p}}$ of $\phi$ is with fine-grained high-level semantic information.

\subsection{Dual-Branch Architecture}
\label{subsec:dual_stream}

During inference, 
we actually can not acquire the text information (e.g., action name) when the network is conducting classification on novel unseen action classes, which means the cross-modal guidance branch can not be used for testing. To solve the asymmetry issue between the training and inference phases, we design a dual-branch architecture that lets the features from both the skeleton encoding branch and the cross-modal guidance branch go through a shared classifier, for action classification during training. 
Such design not only lets features in the skeleton branch learn from the guidance branch during training, but also keeps those two-branch features consistent so that the guidance branch can be removed during inference.


Specifically, after obtaining the fusion feature $p_{st}^{M}$ with both skeleton and textual knowledge, we project it back into the skeleton feature space and pool it along the joint dimension to obtain $f^c_{out}\in \mathbb{R}^{C_{post}}$ as the final output of the cross-modal guidance branch. Similarly, $f_{post} \in \mathbb{R}^{T \times V\times C_{post}}$ in the skeleton encoding branch is also projected and pooled to the same feature space, resulting in $f^s_{out}\in \mathbb{R}^{C_{post}}$. We take $f^s_{out}$ as the final output of the skeleton encoding branch. 
We pass the outputs $f^c_{out}$ and $f^s_{out}$ of the two branches separately through a shared MLP classifier with a softmax layer to obtain the normalized probabilities $\hat{y}_c$ and  $\hat{y}_s$. After that, we use the standard cross-entropy loss to compute the classification losses $\mathcal{L}_{c}$ and $\mathcal{L}_{s}$ of the two branches, respectively, 
\begin{gather}
    \mathcal{L}_{c} = \mathcal{L}_{CE}(\hat{y}_{c},y), \label{eq:l_phi}\\
    \mathcal{L}_{s} = \mathcal{L}_{CE}(\hat{y}_{s},y),  \label{eq:l_cls}
\end{gather}
where y is the action label.
In view of optimization,  the gradient of the cross-modal guidance branch could be back-propagated into the encoding blocks. Thus, the fine-grained high-level semantic information in the joint motion text is passed into the encoding blocks. The enhancement of the encoding blocks can be seen in \cref{fig:spatial_attention}. In the inference phase, we only need the skeleton encoding branch that already incorporates human knowledge for skeleton action recognition, without using any auxiliary text information. Such design also makes our model more efficient during inference, because only the skeleton encoder will be used.

\subsection{Overall Learning Scheme}
\label{subsec:overall_scheme}
The overall training loss of the network is shown in \cref{eq:overall-loss}, where $\alpha _1$ and $\alpha _2$ are the hyperparameters of $\mathcal{L}_{calibrate}$ and $\mathcal{L}_{c}$, respectively:
\begin{equation}
  \mathcal{L}_{overall} = \mathcal{L}_{s} + \alpha _1\mathcal{L}_{calibrate} + \alpha _2\mathcal{L}_{c}.
  \label{eq:overall-loss}
\end{equation} 

At the end of the training, we freeze the skeleton encoding branch and use the distribution calibration method~\cite{DC} for one-shot action classification.

\section{Experiment}
\label{sec:experiments}

To demonstrate the advantages of CrossGLG, we perform one-shot skeleton action recognition on three large-scale datasets, NTU RGB+D 60~\cite{ntu60}, NTU RGB+D 120~\cite{ntu120} and Kinetics~\cite{kinetics}. We compare our model with current SOTA methods and conduct ablation studies to verify the effect of each component.
\subsection{Datasets and Settings}
\noindent{\textbf{NTU RGB+D Dataset 60 (NTU 60)}}~\cite{ntu60} is a large-scale dataset, which contains $56,578$ videos with $60$ action labels and $25$ joints for each body, including interactions with individual and pairs activities. It has a total of $10$ categories of novel actions in the one-shot setting.

\noindent{\textbf{NTU RGB+D Dataset 120 (NTU 120)}}~\cite{ntu120} represents the most extensive dataset for action recognition, comprising $114,480$ videos annotated with $120$ action labels. Recorded across diverse settings, involving $106$ subjects and utilizing $32$ distinct setups, the dataset includes a total of $20$ categories of novel actions within the one-shot setting.

\noindent{\textbf{Kinetics~\cite{kinetics}.} Besides the aforementioned two widely used benchmarks~\cite{ntu120, ntu60}, we further conduct experiment on the challenging Kinetics dataset, with a fair 1-shot setting. It contains around $650,000$ video clips retrieved from YouTube. The videos cover as many as $400$ human action classes, ranging from daily activities, sports scenes, to complex actions with interactions. Please see the appendix for its evaluation protocol under the 1-shot setting.}

\noindent{\textbf{Implementation details.}} 
Please refer to the appendix for the settings of the skeleton encoding branch.
For the cross-modality guidance branch, 3 interaction blocks are used for fusing skeleton and texture features. The hyperparameters $\alpha_1$ and $\alpha_2$ illustrated in \cref{subsec:overall_scheme} are $0.5$ and $0.2$, respectively. DeBERTa-V2-Xlarge~\cite{deberta} is chosen as the text encoder. Two NVIDIA RTX $3090$ GPUs are used for training and testing. 
\begin{table}[!t]
\scriptsize
  \centering
  \begin{tabular}{@{}l|ccccc|c@{}}
    \toprule
    \#Base Classes & 20 & 40 &60 & 80& 100&Para(M)\\
    \midrule
    APSR~\cite{ntu120}           & 29.1  & 34.8  & 39.2  & 42.8  & 45.3 &- \\
    uDTW~\cite{uDTW}          & 32.2  & 39.0  & 41.2  & 45.3    & 49.0  &-\\
    SL-DML~\cite{SLDML}         & 36.7  & 42.4  & 49.0    & 46.4  & 50.9 &11.8\\
    Skeleton-DML~\cite{skeletondml}  & 28.6  & 37.5  & 48.6  & 48.0    & 54.2 &11.8\\
    ALCA-GCN~\cite{ALCAGCN}       & 38.7  & 46.6  & 51.0  & 53.7  & 57.6 &-   \\  \hline
    MotionBERT~\cite{motionbert}    & 35.5 & 54.3 & 56.5 & 52.8  & 61.0 & 60.3  \\
    \rowcolor{mygray}MotionBERT~\cite{motionbert}+CrossGLG  & \textbf{51.0}($\uparrow$15.8) & 56.9($\uparrow$2.6) & 58.4($\uparrow$1.9) &  54.2($\uparrow$1.4) & 62.5($\uparrow$1.5) & 60.8(+0.5)  \\
    HDGCN~\cite{HDGCN}       & 39.0    & 50.4 & 55.8 & 49.8 & 51.9 &1.7  \\
    \rowcolor{mygray}HDGCN~\cite{HDGCN}+CrossGLG   & 43.0($\uparrow$4.0) & 56.7($\uparrow$6.3) & 57.4($\uparrow$1.6)   &55.9($\uparrow$6.1) & 56.8($\uparrow$4.9)&1.8(+0.1)\\
    InfoGCN~\cite{infogcn}       & 37.0    & 53.9 & 58.8 & 55.7 & 56.1  &1.6\\
    InfoGCN~\cite{infogcn}+GAP~\cite{GAP} & 35.06    & 54.8 & 50.82 & 53.23 & 59.9  &1.6 \\
    \rowcolor{mygray}InfoGCN~\cite{infogcn}+CrossGLG          & 45.3($\uparrow$8.3) & \textbf{56.8}($\uparrow$2.9) & \textbf{62.1}($\uparrow$3.3)  & \textbf{61.6}($\uparrow$5.9) & \textbf{62.6}($\uparrow$6.5)&1.7(+0.1)\\
    \bottomrule
  \end{tabular}
  \caption{Comparison of classification accuracy (\%) of different methods for one-shot action recognition on NTU 120~\cite{ntu120}.}
  \vspace{-6mm}
  \label{tab:NTU120_compare_other}
\end{table}
\subsection{Comparison with state-of-the-art methods}
\label{comparison}
\noindent{\textbf{NTU 120.}} From~\cref{tab:NTU120_compare_other}, it can be seen that CrossGLG consistently outperforms the current SOTA methods in large margins, within all experimental settings with different numbers of base classes on NTU 120~\cite{ntu120}. CrossGLG also only brings a neglectable cost (0.1M) compared with the based skeleton encoder~\cite{infogcn}, making it much more efficient than the previous SOTA MotionBERT~\cite{motionbert} (only 2.8\% of its model size) during inference. Besides, we also apply GAP~\cite{GAP}, the current SOTA text-guided method designed for fully supervised setting, under one-shot learning setting. It can be seen that when generalizing to the one-shot learning setting, its performance is even lower than the base encoder InfoGCN~\cite{infogcn} in some cases. Meanwhile, CrossGLG can consistently enhance the performance of the based skeleton encoder with neglectable cost, showing the strong superiority of the proposed method. 
Actually, CrossGLG can be regarded as a plug-and-play module that enhances the feature learning of different existing skeleton encoders (e.g., MotionBERT~\cite{motionbert}, HDGCN~\cite{HDGCN}, and InfoGCN~\cite{infogcn}). All we need to do is change the base skeleton encoder in our skeleton encoding branch.
To verify its effectiveness, we conducted experiments on NTU120~\cite{ntu120} and applied CrossGLG to MotionBERT~\cite{motionbert}, HDGCN~\cite{HDGCN}, and InfoGCN~\cite{infogcn}, respectively. It can be seen that CrossGLG delivers substantial improvements in all settings. Moreover, CrossGLG only brings a neglectable cost compared with the original skeleton encoders, thanks to our dual-branch design. This proves that CrossGLG can be used as a plug-and-play module with good generalization capability.

\noindent{\textbf{NTU 60.}} We also compare SOTA methods with CrossGLG on NTU 60 dataset ~\cite{ntu60}. As in~\cref{tab:NTU60_compare_other}, CrossGLG still outperforms the most SOTA method~\cite{motionbert} in the vast majority of settings. 
Although our method exhibits a 0.4\% lower performance compared to MotionBERT~\cite{motionbert} in the setting of 10 base classes, we attribute this to the relatively weaker base encoder~\cite{infogcn} we employed, and as shown in the bottom two lines of~\cref{tab:NTU60_compare_other}, CrossGLG actually brings a significant performance improvement ($6.8$\%) to the base encoder, showing its effectiveness.

\noindent{\textbf{Kinetics.}} We also compare SOTA methods with CrossGLG on the Kinetics dataset ~\cite{kinetics}. From~\cref{tab:kinetics_compare_other}, it can be observe that even in more complex and challenging environments, our method still outperforms the current SOTA method~\cite{motionbert}. This further illustrates the superiority and generalization ability of our method in skeleton-based one-shot action recognition.

\begin{table}[t]
  \centering
  \scriptsize
   \begin{minipage}{0.52\linewidth}
   \centering
    \scriptsize
  \begin{tabular}{@{}l|ccccc@{}}
    \toprule
    \#Base Classes   & 10   & 20         &30     & 40            & 50\\
    \midrule
    uDTW~\cite{uDTW}           & 56.9 & 61.2       & 64.8  & 68.3          & 72.4 \\
    MotionBERT~\cite{motionbert}     & \textbf{58.3} & 61.0      & \underline{70.0} & \underline{70.3}          & \underline{74.5} \\ 
    InfoGCN~\cite{infogcn}  & 51.1     &\underline{62.1}      & 65.7 & \underline{72.1}  & 72.3 \\   
    \rowcolor{mygray}InfoGCN+CrossGLG  & \underline{57.9}     &\textbf{ 67.1}      & \textbf{70.9} & \textbf{73.4}  & \textbf{75.6} \\
    \bottomrule
  \end{tabular}
  \caption{Comparison of classification accuracy (\%) of different methods for one-shot action recognition on NTU 60~\cite{ntu60}.}
  \label{tab:NTU60_compare_other}
 \end{minipage}
 \hspace{4mm}
 \begin{minipage}{0.4\linewidth}
    \scriptsize
    \centering
      \begin{tabular}{l|cc}
        \toprule
        \#Base Classes & 20   & 40   \\
        \midrule
        MotionBERT~\cite{motionbert}                       & 13.2 & 16.6 \\ 
        InfoGCN~\cite{infogcn}              & \underline{13.3} & \underline{18.2} \\
        \rowcolor{mygray}InfoGCN~\cite{infogcn}+CrossGLG       & \textbf{17.4} & \textbf{19.2} \\
        \bottomrule
        \end{tabular}
        \caption{Comparison of classification accuracy (\%) of different methods for one-shot action recognition on Kinetics~\cite{kinetics}.}
        \label{tab:kinetics_compare_other}
 \end{minipage}
 \hspace{-1mm}
 \vspace{-25pt}
\end{table}


\subsection{Ablation Studies}
Here we choose InfoGCN~\cite{infogcn} as the base skeleton encoder and perform ablation studies on the NTU 120~\cite{ntu120} to analyze the role of the model components.



\noindent\textbf{Analysis of global-to-local textual guidance (G2L).}
The result is shown in~\cref{tab:ablation_study}. It can be observed that G2L can consistently improve one-shot action recognition across all settings, which essentially demonstrates the effectiveness of the proposed global-to-local textual guidance. 
We also experimented with the insertion location of the proposed JID module. The encoder (InfoGCN~\cite{infogcn}) has a total of $9$ encoding blocks. JID takes the output of the $N_{pre}$-th block as input. For example, when $N_{pre}$ is $9$, it means that the output of JID is directly applied to the output of the $9$-th layer. The results of our experiments for the setting of $N_{pre}$ are shown in \cref{tab:phase1_layers}. It can be seen that the overall classification accuracy of the model is best when the output of the $5$-th block is fed to JID. Our analysis is that when the insertion position of the JID is too forward, the skeleton features fed into the JID are not sufficiently fused with spatio-temporal information, and it is difficult for the JID to effectively determine the importance of each joint through such shallow features. On the other hand, when the insertion position is too far back, the subsequent blocks that can be affected by the JID are too few so it is difficult to influence the feature representations.


\begin{table}[t]
  \centering
  \scriptsize
   \begin{minipage}{0.45\linewidth}
   \centering
    \scriptsize
    
      \begin{tabular}{@{}cc|ccccc@{}}
        \toprule
        \multirow{2}{*}{G2L}    & \multirow{2}{*}{L2G} & \multicolumn{5}{c}{\#Base classes}    \\ \cline{3-7} 
                                &                      & 20    & 40    & 60    & 80    & 100   \\ \hline
        ×                                       & ×    & 37    & 53.9 & 58.8 & 55.7 & 56.1 \\
        \checkmark               & ×         & 43.3 & 54.7 & 60.9 & 58.7 & 61.7 \\
        ×                        & \checkmark    & 42.5 & 54.9 & 61.7  & 60.1 & 58.6 \\
        \checkmark               & \checkmark   & \textbf{45.3} & \textbf{56.8} & \textbf{62.1}  & \textbf{61.6} & \textbf{62.6} \\
        \bottomrule
      \end{tabular}
      \caption{Performance comparison with and without the proposed G2L and L2G on the NTU 120~\cite{ntu120}.}
      \label{tab:ablation_study}
 \end{minipage}
 \hspace{5mm}
 \begin{minipage}{0.45\linewidth}
    \scriptsize
    \centering
  \begin{tabular}{@{}cccccc@{}}
    \toprule
    \multicolumn{1}{c}{\multirow{2}{*}{$N_{pre}$}} & \multicolumn{5}{c}{\#Base Classes}                               \\ \cline{2-6} 
    \multicolumn{1}{c}{}                      & 20         & 40         & 60         & 80         & 100        \\ \hline
    3                                         & 41.1 & 50.8 & \textbf{61.9} & 54.5 & 57.7 \\
    5                                         & \textbf{43.3}      & \textbf{54.7}      & 60.9   & \textbf{58.7}   & \textbf{61.7}      \\
    7                                         & 43.2 & 54.1 & 58.7              & 58.5      & 58.0      \\
    9                                         & 38.7 & 51.8 & 54.4              & 57.3 & 55.3 \\
    \bottomrule
  \end{tabular}
  \caption{Performance Analysis of Insertion Position $N_{pre}$ for JID Module on NTU 120~\cite{ntu120}.}
  \label{tab:phase1_layers}
 \end{minipage}
 \hspace{-5mm}
 \vspace{-17pt}
\end{table}

\noindent\textbf{Analysis of local-to-global cross-modal interaction (L2G).}
 As can be observed from \cref{tab:ablation_study}, when L2G is added, the classification accuracy under different experimental settings is also improved in all cases. This proves that the proposed L2G successfully conveys rich high-level semantic context to guide the skeleton feature learning. 
 We also conducted experiments on the text-text interaction within the interaction block $\phi$. As shown in \cref{tab:txt_self_attention}, it brings consistent performance gain across all settings. We argue this is because it can summarize non-local semantics to facilitate the subsequent skeleton-text interaction.
 
 


\begin{table}[t]
  \centering
  \scriptsize
   \begin{minipage}{0.58\linewidth}
   \centering
    \scriptsize
  \scalebox{0.9}{ \begin{tabular}{@{}cccccc@{}}
    \toprule
    Text-Text & \multicolumn{5}{c}{\#Base classes}    \\ \cline{2-6} 
     interaction                                     & 20    & 40    & 60    & 80    & 100   \\ \hline
    ×                                                        & 41.7 & 53.8 & 59.5 & 49.6 & 55.9 \\
    \checkmark                                                        & \textbf{42.5} & \textbf{54.9} & \textbf{61.7}  & \textbf{60.1} & \textbf{58.6} \\
    \bottomrule
  \end{tabular}}
  \caption{Analysis of the role of the text self-attention module in the interaction block of joint skeleton features and joint textual features interaction module $\phi$ on NTU 120~\cite{ntu120}.}
  \label{tab:txt_self_attention}
 \end{minipage}
 \hspace{4mm}
 \begin{minipage}{0.36\linewidth}
    \scriptsize
    \centering
    \vspace{-1mm}
\scalebox{0.9}{ \begin{tabular}{lc}
\toprule
LLM     & Acc(\%)    \\
\midrule
ChatGPT~\cite{GPT3} & 45.3   \\
Gemini~\cite{gemini}   & 45.2   \\
Qwen~\cite{Qwen}      &45.7    \\
\bottomrule
\end{tabular}}
\caption{Robustness analysis of CrossGLG under different LLM on NTU 120~\cite{ntu120} with 20 base classes.}
\label{tab:LLM}
 \end{minipage}
 \vspace{-30pt}
\end{table}


\noindent\textbf{Robustness to LLMs.} Here we evaluate the robustness of CrossGLG when using different LLM for text generation. As shown in~\cref{tab:LLM}, CrossGLG is robust with different LLM (performance difference less that 0.5\%).

\section{Visual Analysis}
\label{sec:analysis}
We visualize the intermediate results of the network to further analyze the effectiveness of the proposed method. The experiments are conducted on the NTU 120~\cite{ntu120} dataset with a base class of $20$ classes.



\begin{figure}[t]
  \centering
   \includegraphics[width=0.8\linewidth]{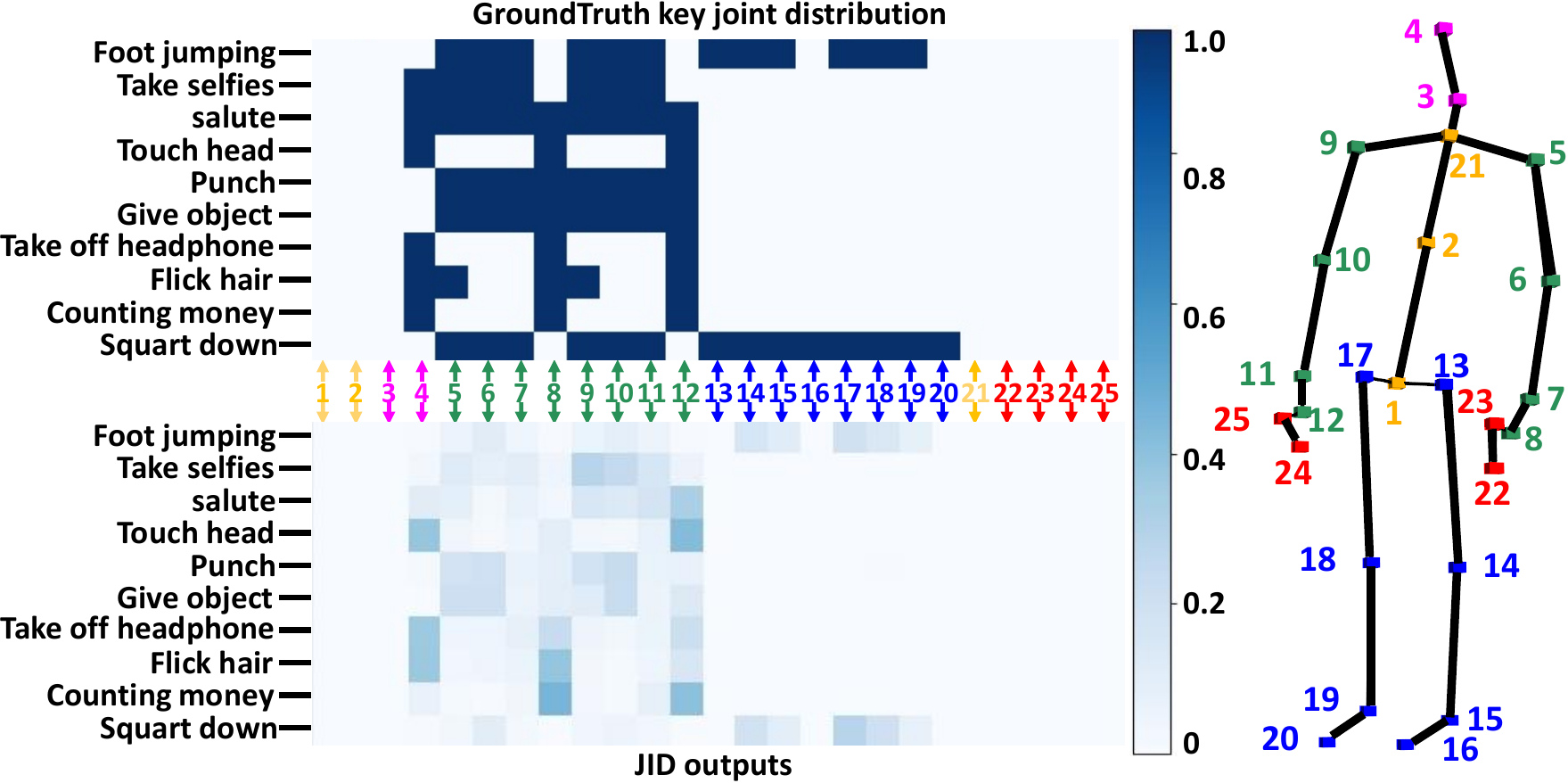}
   \caption{Visualization of the output joint importance from JID. The upper half of the distribution of Ground Truth key joints is extracted from the global action description. The bottom half is the output of the Joint Importance Discrimination (JID) module.}
   \label{fig:JID_out}
   \vspace{-5mm}
\end{figure}

\noindent\textbf{Visualization of the output joint importance from JID.} We visualize the output of JID (the bottom part of ~\cref{fig:JID_out}) and compare it with the ground truth key joint distribution (the upper part of ~\cref{fig:JID_out}). It can be seen that JID successfully learned to predict joint importance which is similar to the GT key joint distribution. At the same time, JID does not simply fit the GT distribution, but also adjusts the weights appropriately according to the observed skeleton features. For example, the GT key joint of `Squat down' derived from the global action description contains joints around both leg and arm. However, the movement of arm is actually not strongly correlated with the action of squatting down. This is also learned by JID, as it produces low weights on joints around arms, which further demonstrates its capacity for high-level semantic understanding.



\begin{figure}[t]
  \centering
   \includegraphics[width=0.8\linewidth]{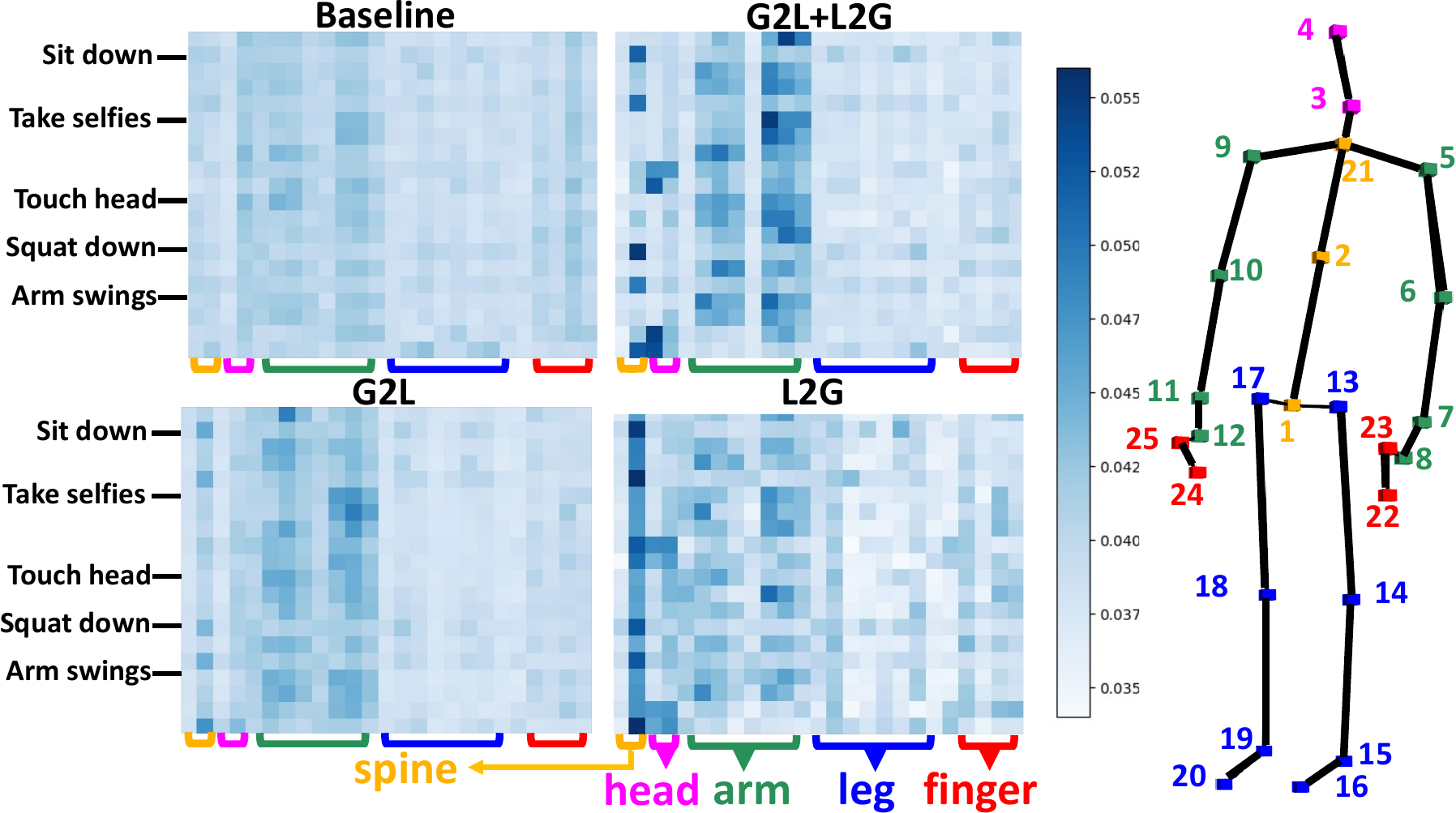}
   \caption{Visual analysis of spatial attention within the skeleton encoder. In each matrix, each row represents the importance of all encoding blocks for each joint feature in an action. The deeper the color, the higher the importance.}
   \label{fig:spatial_attention}
\end{figure}

\noindent\textbf{Attention visualization of the skeleton encoder.}
We visualize the attention of the skeleton encoder to analyze its behaviors, which is shown in \cref{fig:spatial_attention}. In each visualization matrix, each row represents the spatial attention of each joint in an action. The color assigned to each joint on the right side in \cref{fig:spatial_attention} corresponds to the color representing its respective position at the bottom of the matrix. It can be seen that G2L, L2G, and the combination of them significantly improve the spatial clue-grasping ability of the model. For example, compared to the baseline, they improved the attention for the arm and finger parts in both `touch head' and `arm swings'. In the actions `sit down' and `squat down', there is more spatial attention to the leg joints. 



\begin{figure}[t]
  \centering
   \includegraphics[width=0.95\linewidth]{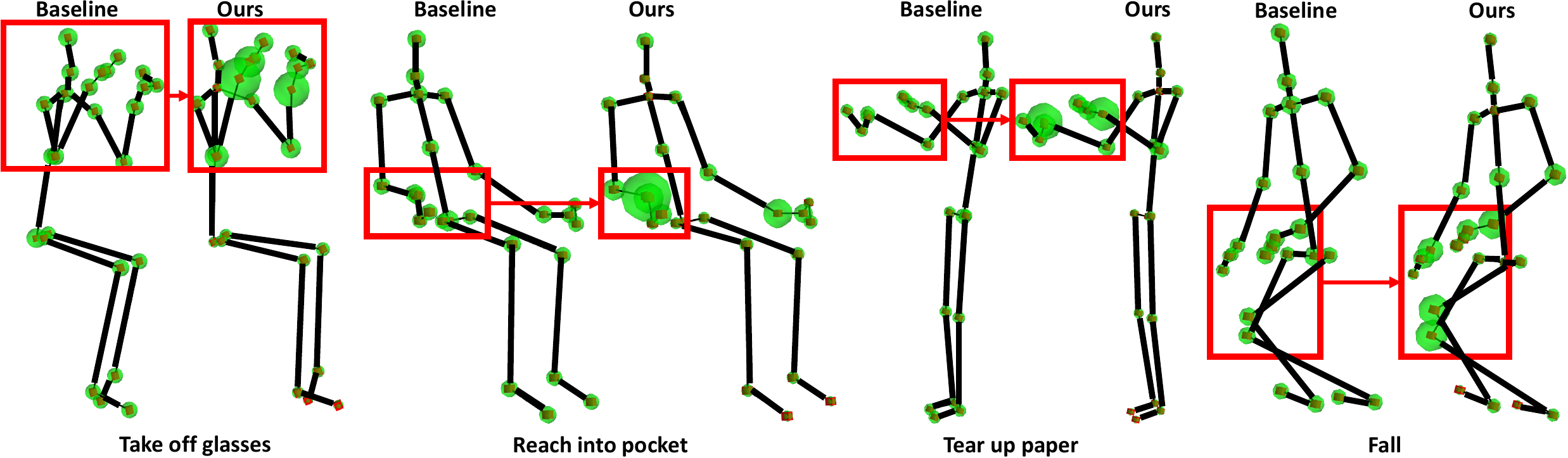}

   \caption{Spatial attention visualization analysis on novel class actions. The spheres around each joint indicate how much attention the model pays to it. We draw the `Take off glasses', `Peach into pocket', `Tear up paper' and `Fall' actions for illustration.}
   \label{fig:novel_class}
   \vspace{-6mm}
\end{figure}

We also perform a visual analysis of the spatial attention of the novel class actions. At the end of the training of the model, we freeze the skeleton encoding branch and directly examine the spatial attention of the novel actions, and the results are shown in \cref{fig:novel_class}. It can be seen that compared to the baseline, our method pays more attention to the hand features in the actions `Take off glasses' and `Tear up paper'. In the action `reach into pocket', our method pays more attention to the right hand that reaches into the pocket. In the `Fall' action, our model not only focuses on key joints such as the knees and hands but also pays attention to areas like the spine, as the fall is an action relevant to the whole body. The ability of key spatial information capturing in unseen action classes demonstrates the strong generalization of our method.

\section{Conclusion}
\label{sec:conclusion}
In this work, we propose a novel cross-modal guidance framework CrossGLG that leverages action descriptions rich in human knowledge to guide the skeleton feature learning, in a global-local-global way. Specifically, the proposed global-to-local guidance enables the skeleton encoder to focus more on important local information. The proposed local-to-global guidance, on the other hand, establishes a non-local interaction between joint-level motion descriptions and skeleton features to guide the generation of global action representations using fine-grained high-level semantic information. In addition, we design a dual-branch architecture to mitigate the asymmetry issue between the training and inference phases. Experimental results verify the superiority of CrossGLG in effectiveness, efficiency, and strong capacity as a plug-and-play module.

\clearpage  

\setcounter{section}{0}
\setcounter{table}{0}
\setcounter{figure}{0}
\renewcommand{\thetable}{\AA \arabic{table}}
\renewcommand{\thefigure}{\AA \arabic{figure}}
\renewcommand{\thesection}{\AA \arabic{section}}
\begin{center}
    \textbf{\huge {Appendix}}
\end{center}
\section{Implementation details}
InfoGCNs~\cite{infogcn} was mainly chosen as the skeleton coding branch for our experiment. During training, the learning rate is set to $0.05$, batch\_size is set to $128$, and the number of training epochs is $110$. $\alpha _1$ and $\alpha _2$ are set to $0.5$ and $0.2$, respectively. The random seed is $0$. During testing, the batch\_size is also set to $128$. When applying GAP~\cite{GAP} to the 1-shot task, as described in its paper, the learning rate is set to $0.1$, and the trade-off parameter $\lambda$ is set to $0.8$. Additionally, in the experiments of \cref{comparison}, the learning rate for HDGCN~\cite{HDGCN} is set to $0.05$. The learning rate for MotionBERT's~\cite{motionbert} backbone is set to $0.00008$. $\alpha_1$ and $\alpha_2$ are also set to $0.5$ and $0.2$, respectively, while other settings remain unchanged from their initial configurations. We use Pytorch to implement our work. All models in our experiments were done on two NVIDIA RTX $3090$ GPUs.

\section{Training and evaluation protocol on Kinetics}
We followed the skeleton extraction as well as the processing method in ~\cite{STGCN} and sampled each skeleton sequence evenly over 60 frames. Since kinetics~\cite{kinetics} did not have an official 1-shot setting before, we designed a fair base classes set and novel classes set split following NTU120~\cite{ntu120}. As described in \cref{subsec:preliminary}, the model will be trained using a $\mathcal{D}_{base}$ consisting of all the samples from the training set. One sample from each of the Novel classes will be randomly selected to form a support set $\mathcal{S}_{novel}$ to help the model categorize the unseen samples from the Novel classes. The support set $\mathcal{S}_{novel}$ is kept constant in the same setting. The following are the splits for the 20 base classes and the 40 base classes, respectively:
\begin{itemize}
    \item \textbf{20-class:}
    \begin{itemize}
        \item \textbf{Train:}  $[$ 2, 22, 42, 62, 82, 102, 122, 142, 162, 182, 202, 222, 242, 262, 282, 302, 322, 342, 362, 382$]$ 
        \item \textbf{Novel:} $[$ 3, 23, 43, 63, 83, 103, 123, 143, 163, 183, 203, 223, 243, 263, 283, 303, 323, 343, 363, 383 $]$
    \end{itemize}
    
    \item \textbf{40-class:}
    \begin{itemize}
        \item \textbf{Train:} $[$ 2,   4,  22,  24,  42,  44,  62,  64,  82,  84, 102, 104, 122, 124, 142, 144, 162, 164,
     182, 184, 202, 204, 222, 224, 242, 244, 262, 264, 282, 284, 302, 304, 322, 324, 342, 344,
     362, 364, 382, 384 $]$
        \item \textbf{Novel:} $[$ 3, 23, 43, 63, 83, 103, 123, 143, 163, 183, 203, 223, 243, 263, 283, 303, 323, 343, 363, 383 $]$
    \end{itemize}
\end{itemize}

\section{The impact of hyperparameters $\alpha _1$ and $\alpha _2$}
\begin{table}[]
  \centering
    \begin{tabular}{cccc|cccc}
    \toprule
    \multicolumn{4}{c|}{$\alpha_1$}     & \multicolumn{4}{c}{$\alpha_2$}     \\
    \midrule
    0    & 0.2  & 0.5  & 0.6  & 0    & 0.2  & 0.3  & 0.5  \\
    42.5 & 43,6 & 45.3 & 44.9 & 43.3 & 45.3 & 44.8 & 45.1 \\
    \bottomrule
    \end{tabular}
  \caption{Choice for $\alpha_1$ and $\alpha_2$ at 20 base classes settings on NTU120~\cite{ntu120}.}
  \label{tab:NTU120_a1a2}
\end{table}

$\alpha _1$ and $\alpha _2$ are used to balance the overall effect of the proposed losses $\mathcal{L}_{calibrate}$ and $\mathcal{L}_{c}$. \cref{tab:NTU120_a1a2} shows that the performance is relatively robust to their values.

\section{Limitations}
CrossGLG realizes the introduction of human knowledge in text from two perspectives, global-to-local and local-to-global, respectively. Although results from different experimental settings have demonstrated the effectiveness of CrossGLG, it still has some room for improvement in some extremely difficult-to-distinguish situations. For example, there are still a certain number of misjudgments in the actions `yawn' and `hush'.

\section{Visual analysis}
\begin{figure}[t]
  \centering
   \includegraphics[width=1\linewidth]{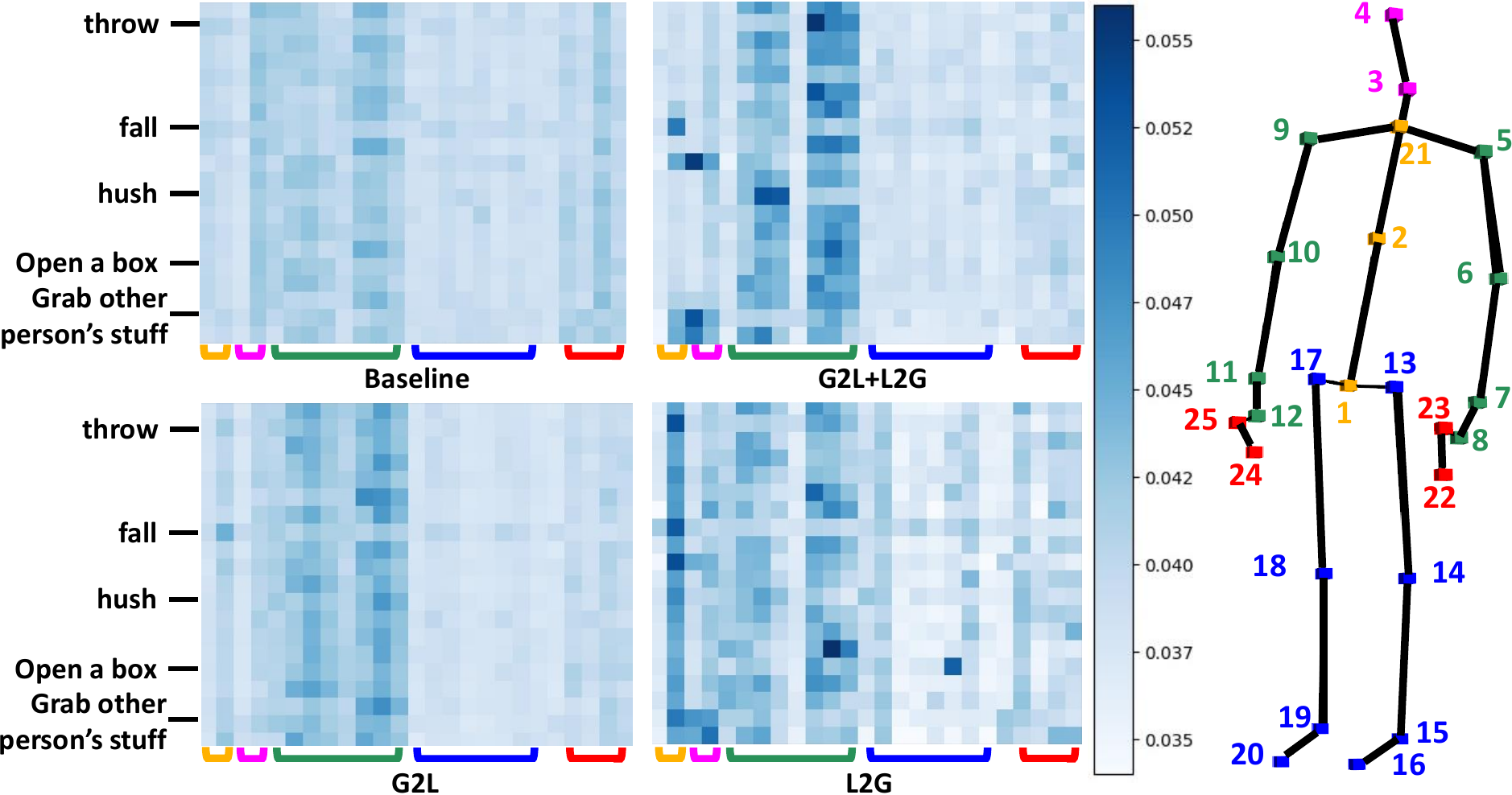}

   \caption{Visual analysis of spatial attention within the skeleton encoder. In each matrix, each row represents the importance of all encoding blocks for each joint feature in a novel action. The deeper the color, the higher the importance.}
   \label{fig:novel_attention}
\end{figure}
We visualize the attention of the skeleton encoder to analyze its behaviors, which is shown in \cref{fig:novel_attention}. In each visualization matrix, each row represents the spatial attention of each joint in a novel action. The color assigned to each joint on the right side in \cref{fig:novel_attention} corresponds to the color representing its respective position at the bottom of the matrix. At the end of the training of the model, we freeze the skeleton encoding branch and directly examine the spatial attention of the novel actions. G2L, L2G, and a combination of the two can focus more on the really important joints in the actions than the baseline. For example, in the actions `throw' and `open a box', the arms and hands are more important. In the action `fall', which involves the joints of the whole body, the network gives similar importance to each joint according to the characteristics of the action, meaning that the network pays attention to all joints.

\section{Text descriptions}
We use large language models (such as ChatGPT~\cite{GPT3}) to obtain the global action description and joint motion descriptions for each action.
The global action description indicates which joints and body parts are important from a global perspective. The joint-level motion description provides fine-grained high-level semantic information from a local perspective. The global motion descriptions and joint motion descriptions for several actions are shown below:

\begin{center}
\centering
\textbf{Wave hand}
\end{center}
\begin{itemize}
    \item \textbf{Global action description:} one \textbf{arm} is raised and moved side to side, typically with the \textbf{hand} open, often performed in a rhythmic motion, involving a swinging movement at the \textbf{elbow} joint while the \textbf{wrist} creates a slight flexion and extension
    \item \textbf{joint motion descriptions:} 
    \begin{itemize}
        \item \textbf{Base of spine:} Generally remains stable, providing a balanced foundation for the upper body.
        \item \textbf{Mid of spine:} Typically straight and upright, supporting the upper body's movements.
        \item \textbf{Neck:} Allows for flexibility and rotation, facilitating head movement.
        \item \textbf{Head:} May tilt slightly to the side or stay upright, often facing the direction of the hand waving.
        \item \textbf{Left shoulder:} Lifts and rotates slightly as the arm moves to wave.
        \item \textbf{Left elbow:} Bends and straightens rhythmically to create the waving motion.
        \item \textbf{Left wrist:} Flexes and extends, contributing to the waving movement.
        \item \textbf{Left hand:} Performs a waving motion, lifting and lowering in a friendly gesture.
        \item \textbf{Right shoulder:} Remains engaged in supporting the arm's movement.
        \item \textbf{Right elbow:} Bends and straightens in coordination with the waving action.
        \item \textbf{Right wrist:} Flexes and extends, mirroring the movements of the left wrist.
        \item \textbf{Right hand:} Performs the waving action, lifting and lowering in a friendly gesture.
        \item \textbf{Left hip:} Typically stable during hand waving, but may sway slightly with body movement.
        \item \textbf{Left knee:} Generally remains straight but can bend slightly with body weight shifts.
        \item \textbf{Left ankle:} Often stationary or may have subtle movements with body weight shifts.
        \item \textbf{Left foot:} Remains grounded, providing stability.
        \item \textbf{Right hip:} Similar to the left hip, usually stable during hand waving.
        \item \textbf{Right knee:} Generally straight but may bend slightly with body weight shifts.
        \item \textbf{Right ankle:} Often stationary or may have subtle movements with body weight shifts.
        \item \textbf{Right foot:} Remains grounded, providing stability.
        \item \textbf{Spine:} Supports an upright posture, allowing for coordination of upper body movements.
        \item \textbf{Tip of Left hand:} Moves in a waving motion, extending and retracting.
        \item \textbf{Left thumb:} Provides stability during the waving action.
        \item \textbf{Tip of Right hand:} Mimics the waving motion of the left hand, extending and retracting.
        \item \textbf{Right thumb:} Offers stability and support during the waving action.
    \end{itemize}
\end{itemize}

\begin{center}
\centering
\textbf{Squat down}
\end{center}
\begin{itemize}
    \item \textbf{Global action description:} the \textbf{knees} and \textbf{hips} bend while the torso lowers, maintaining an upright posture, and the \textbf{arms} may extend forward or rest on the thighs, with the body weight shifting towards the heels, engaging the \textbf{leg} muscles for support and balance.
    \item  \textbf{joint motion descriptions:}
    \begin{itemize}
        \item \textbf{Base of spine:} The base of the spine is actively engaged in the action, supporting the weight of the upper body as it moves downward.
        \item \textbf{Mid of spine:} The mid of the spine also participates by flexing forward slightly as the upper body leans forward during the squat.
        \item \textbf{Neck:} The neck remains in line with the spine, maintaining a neutral position.
        \item \textbf{Head:} The head aligns with the spine and does not actively move during the squat.
        \item \textbf{Left shoulder:} The left shoulder remains stationary, with no significant movement.
        \item \textbf{Left elbow:} The left elbow may slightly flex as the left arm moves with the body during the squat.
        \item \textbf{Left wrist:} The left wrist maintains its natural position, without significant movement.
        \item \textbf{Left hand:} The left hand is relaxed and usually hangs down as the body squats.
        \item \textbf{Right shoulder:} Similar to the left shoulder, the right shoulder remains stationary.
        \item \textbf{Right elbow:} The right elbow may slightly flex as the right arm moves with the body during the squat.
        \item \textbf{Right wrist:} The right wrist maintains its natural position, without significant movement.
        \item \textbf{Right hand:} The right hand is relaxed and usually hangs down during the squat.
        \item \textbf{Left hip:} The left hip is actively involved in the squatting motion, allowing the left thigh to move backward.
        \item \textbf{Left knee:} The left knee flexes as the body lowers, allowing the squatting motion to occur.
        \item \textbf{Left ankle:} The left ankle allows for dorsiflexion, which aids in the squatting movement.
        \item \textbf{Left foot:} The left foot remains in contact with the ground, providing support.
        \item \textbf{Right hip:} Similar to the left hip, the right hip is actively involved in the squatting motion, allowing the right thigh to move backward.
        \item \textbf{Right knee:} The right knee flexes as the body lowers, mirroring the movement of the left knee.
        \item \textbf{Right ankle:} The right ankle allows for dorsiflexion, aiding in the squatting movement.
        \item \textbf{Right foot:} The right foot remains in contact with the ground, providing support.
        \item \textbf{Spine:} The entire spine is involved in the squatting motion, with the base and mid of the spine flexing slightly forward to maintain balance.
        \item \textbf{Tip of Left hand:} The tip of the left hand usually remains relaxed and hangs down during the squat.
        \item \textbf{Left thumb:} The left thumb remains in a relaxed position, with no significant movement.
        \item \textbf{Tip of Right hand:} Similar to the left hand, the tip of the right hand usually remains relaxed and hangs down.
        \item \textbf{Right thumb:} The right thumb remains in a relaxed position, with no significant movement.
    \end{itemize}
\end{itemize}

\begin{center}
\centering
\textbf{Count money}
\end{center}
\begin{itemize}
    \item \textbf{Global action description:} the \textbf{hands} are involved in picking up, sorting, and organizing the bills or coins, while the \textbf{eyes} focus on visually inspecting and verifying the amount, and the \textbf{head} is engaged in mental calculations and cognitive processing to keep track of the counting progress
    \item \textbf{joint motion descriptions:} 
    \begin{itemize}
        \item \textbf{Base of spine:} The base of the spine generally remains stable while sitting or standing.
        \item \textbf{Mid of spine:} The mid of the spine maintains its natural posture, with no significant movement.
        \item \textbf{Neck:} The neck is often upright or slightly bent forward, depending on the position of the money.
        \item \textbf{Head:} The head is typically upright, and the gaze is directed toward the money being counted.
        \item \textbf{Left shoulder:} The left shoulder usually remains stable and doesn't actively move.
        \item \textbf{Left elbow:} The left elbow is bent at an angle to bring the currency closer for counting.
        \item \textbf{Left wrist:} The left wrist is typically flexed as it guides the movement of bills or coins.
        \item \textbf{Left hand:} The left hand holds and manipulates the money, separating and organizing it.
        \item \textbf{Right shoulder:} The right shoulder typically remains stable for balance.
        \item \textbf{Right elbow:} The right elbow may be bent slightly as it assists in the movement of money.
        \item \textbf{Right wrist:} The right wrist is often flexed to help manage the currency's position.
        \item \textbf{Right hand:} The right hand holds and manipulates the money, assisting in counting and organizing.
        \item \textbf{Left hip:} The left hip provides balance and support but doesn't actively move.
        \item \textbf{Left knee:} The left knee may be slightly bent or adjusted for comfort while sitting or standing.
        \item \textbf{Left ankle:} The left ankle generally maintains its neutral position.
        \item \textbf{Left foot:} The left foot is typically stationary during the counting process.
        \item \textbf{Right hip:} The right hip provides balance and support but doesn't actively move.
        \item \textbf{Right knee:} The right knee may be slightly bent or adjusted for comfort while sitting or standing.
        \item \textbf{Right ankle:} The right ankle generally maintains its neutral position.
        \item \textbf{Right foot:} The right foot is typically stationary during the counting.
        \item \textbf{Spine:} The spine retains its natural posture with no significant movement.
        \item \textbf{Tip of Left hand:} The tip of the left hand is actively engaged in handling and counting the money.
        \item \textbf{Left thumb:} The left thumb may assist in separating or flipping bills.
        \item \textbf{Tip of Right hand:} The tip of the right hand is actively engaged in handling and counting the money.
        \item \textbf{Right thumb:} The right thumb may assist in separating or flipping bills.
    \end{itemize}
\end{itemize}

%
%
\bibliographystyle{splncs04}
\bibliography{main}
\end{document}